\documentclass[review]{elsarticle}
\usepackage{url}
\usepackage{color}
\usepackage{booktabs}
\usepackage{multirow}
\usepackage{multicol}
\usepackage{graphicx}
\usepackage{algorithm}
\usepackage{algorithmic}
\usepackage{lineno,amsmath, amsfonts,graphicx,bm}
\usepackage[colorlinks,
            linkcolor=red,
            anchorcolor=blue,
            citecolor=green
            ]{hyperref}
\usepackage[marginal]{footmisc}

\modulolinenumbers[5]

\journal{Journal of Pattern Recognition}

\biboptions{sort&compress}

\begin{document}

\begin{frontmatter}

\title{Leaf Cultivar Identification via Prototype-enhanced Learning}

\author{Yiyi Zhang$^{1}$}
\author{Zhiwen Ying$^{1}$}
\author{Ying Zheng$^{2}$, Cuiling Wu$^{1}$, Nannan Li$^{1}$, Jun Wang$^{1}$, Xianzhong Feng$^{3}$, Xiaogang Xu$^{4}$\corref{mycorrespondingauthor}}
\cortext[mycorrespondingauthor]{Corresponding author}
\address{$^{1}$ Institute of Intelligent computing, Zhejiang Lab, Hangzhou, China\\
$^{2}$ AI Lab, Gientech, Hangzhou, China\\
$^{3}$ Northeast Institute of Geography and Agroecology, Chinese Academy of Sciences, Changchun, China\\
$^{4}$ School of Computer Science and Technology, Zhejiang Gongshang University, Hangzhou, China\\
yiyi.zhang93@outlook.com, zhengying@hit.edu.cn, xxgang2013@163.com}

\begin{abstract}
\footnote{First Author and Second Author contributed equally.} 
Plant leaf identification is crucial for biodiversity protection and conservation and has gradually attracted the attention of academia in recent years. Due to the high similarity among different varieties, leaf cultivar recognition is also considered to be an ultra-fine-grained visual classification (UFGVC) task, which is facing a huge challenge. In practice, an instance may be related to multiple varieties to varying degrees, especially in the UFGVC datasets. However, deep learning methods trained on one-hot labels fail to reflect patterns shared across categories and thus perform poorly on this task. To address this issue, we generate soft targets integrated with inter-class similarity information. Specifically, we continuously update the prototypical features for each category and then capture the similarity scores between instances and prototypes accordingly. Original one-hot labels and the similarity scores are incorporated to yield enhanced labels. Prototype-enhanced soft labels not only contain original one-hot label information, but also introduce rich inter-category semantic association information, thus providing more effective supervision for deep model training. Extensive experimental results on public datasets show that our method can significantly improve the performance on the UFGVC task of leaf cultivar identification.

\end{abstract}


\end{frontmatter}


\section{Introduction}
\label{sec:introduction}
Cultivar identification plays a vital role in the evaluation, breeding, and production of plant multi-variety. In botany, plant leaves are widely utilized \cite{wang2020species} to identify varieties due to their stable, persistent, and detective morphological characteristics. Conventional methods of plant identification often require empirical knowledge that is not readily available to non-experts. The process of manual variety classification is time-consuming and error-prone, making it difficult to meet the high demand for plant ecological research.  

Over the past decade, researchers have successfully extracted topological features from leaves in terms of texture, shape, and venation. However, these hand-crafted traits are low-level features \cite{zheng2019forest}, which are not descriptive enough to distinguish cultivars belonging to the same species. Intuitively, deep learning is ideal to extract rich high-level features from leaf images. Existing leaf classification methods can be roughly divided into two sets. One set fuses multiple features extracted from different parts of leaves or through different manners to jointly identify leaf cultivar \cite{wang2022fusing, yu2021maskcov,zhang2021mfcis,lee2017deep}. These hybrid methods require extra computational cost to calculate a group of inputs. The other set performs cascaded frameworks to encode features through a pipeline \cite{beikmohammadi2022swp,zheng2022fuzzy,yu2023mix}. Nevertheless, these methods need additional network structures and some of them cannot be trained in an end-to-end manner. In brief, existing deep learning methods for leaf cultivar classification are not effective to meet the demands of practical scenarios.

\begin{figure*}[t]
	\centering
	\centerline{\includegraphics[width=11.5cm]{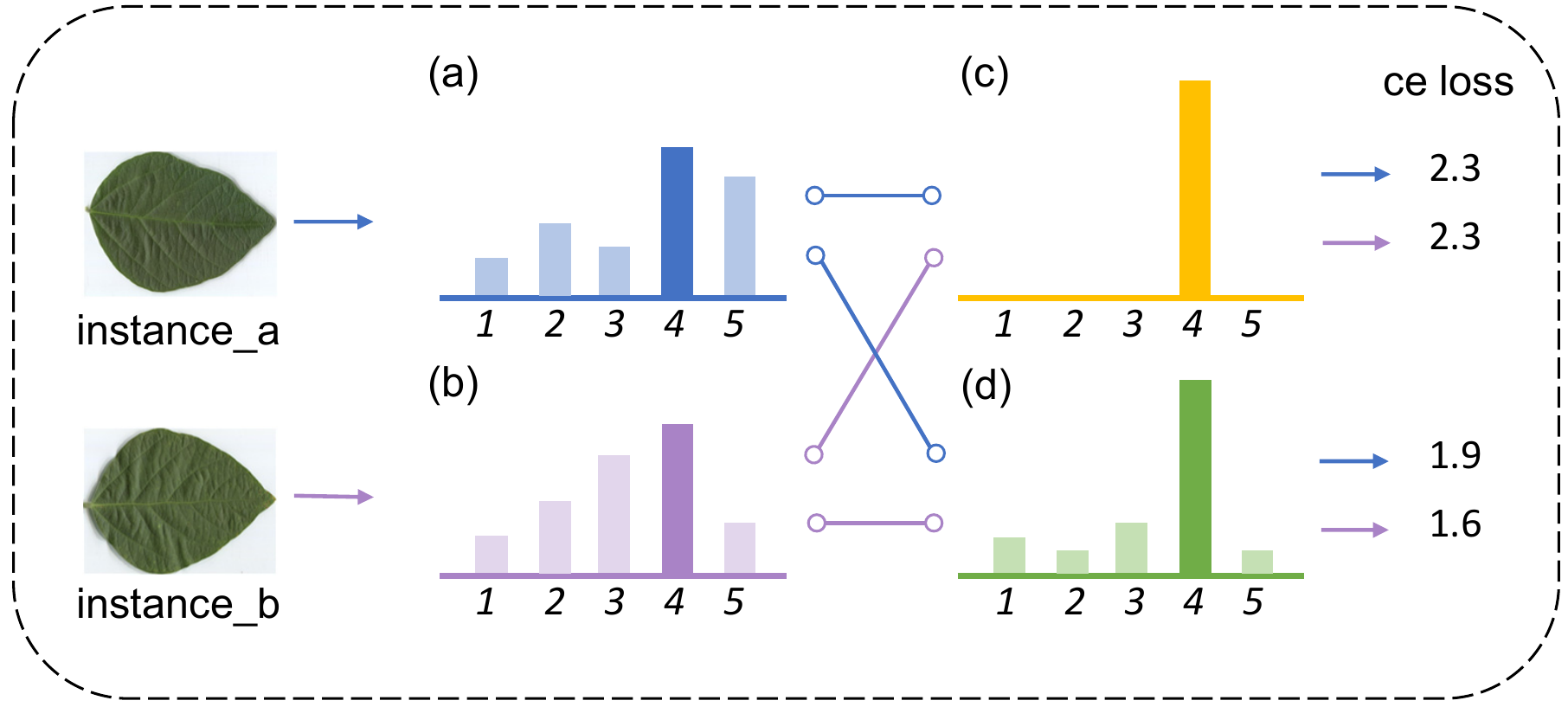}}
	\caption{A case study comparing the effect of the soft and hard targets. (a) and (b) are predicted label distributions of two instances from the same class; (c) is the one-hot hard target; (d) is the simulated soft target. Though both instances are classified to the positive class 4 with the same ratio, negative classes carry different ratios of probability in (a) and (b). By computing the cross-entropy (ce) loss, the soft target reveals this information gap while the hard target treats (a) and (b) equally. $\circ$-$\circ$ represents distance calculation between two distributions. }
	\label{fig:intro}
\end{figure*}

Moreover, deep learning approaches rely heavily on large amounts of annotated data, where the quality of labels is critical to model performance. In many cases, errors in manual labeling are inevitable, directly leading to performance degradation. The mislabeling problem is exacerbated especially in UFGVC datasets where each class is highly similar. Meanwhile, the typical one-hot labeling used to train deep models assigns the full probability to one single class, making the model particularly vulnerable to mislabeled instances and appearing overconfident. As expected, works of knowledge distilling \cite{hinton2015distilling, gou2021knowledge} have claimed the advantages of soft targets. When the soft targets have high entropy, they are much more likely to provide richer information per training case than hard targets (one-hot labels). As illustrated in Figure \ref{fig:intro}, all negative classes are treated equally in hard targets, while soft targets can measure the difference behind negative classes, leading to different gradient directions in the training phase. A label smoothing(LS) method \cite{pereyra2017regularizing} was proposed to alleviate the limitation of one-hot labels through a “confidence penalty”. However, \cite{muller2019does} demonstrated that LS loses information in the logits about resemblances among different classes. In light of these pros and cons, we thus ask: how to generate soft targets for improved ultra-fine-grained classification? 

As an analogy to natural language processing (NLP), by capturing the co-relation between labels, label embedding \cite{yang2018sgm} can select the most informative words and neglect irrelevant ones when predicting different labels for multi-label text classification. Based on this intuition, we propose a novel Prototype-enhanced Learning (PEL) method for leaf cultivar identification. PEL is predicated on the assumption that label embedding encoded with the inter-class relationships would force the image classification model to focus on discriminative patterns. Unlike classical label embedding in NLP, we capture inter-class co-relation at the feature level by generalizing categorical prototypes. The idea of prototypes is related to prototypical networks \cite{snell2017prototypical,ji2020improved} in few-shot learning. They assign instance labels to the closest prototypes in the classifier. However, we use prototypes to augment original labels, which is significantly different from them. In PEL, the prototypes are updated via a moving average during training, thus continuously approaching the corresponding class centers on-the-fly. We conduct similarity scoring between input features and prototypes. Similarity scores are fused into one-hot label representations to generate enhanced labels. We use the obtained logits to replace original labels for supervised model learning. 

With the help of PEL, a deep network not only learns to distinguish varieties but also grasps the semantic relationship between each label. The proposed PEL is compared with 22 state-of-the-art methods including hand-crafted feature descriptors, CNN-based and transformer-based deep learning methods. Encouraging experimental results are reported on 7 ultra-fine-grained image datasets, demonstrating the effectiveness of PEL for the UFGVC tasks.  The main contributions are summarized as follows:
\begin{itemize}
\item A novel method named Prototype-enhanced Learning (PEL) is proposed for leaf cultivar identification. Compared to existing methods, PEL is end-to-end trainable and adds no extra parameters, which is light and effective with negligible computational overhead.
\item We develop a new prototype update module to learn inter-class relations by capturing label semantic overlap and iteratively update prototypes to generate continuously enhanced soft targets.
\item Extensive experiments on 7 benchmarks adequately illustrate the superiority of PEL for UFGVC. State-of-the-art performances are reported on two widely used backbones including ResNet and DenseNet.
 
\end{itemize}

\section{Related Work}
This section reviews various plant leaf recognition methods from the species level and cultivar level separately. In detail, some representative methods are surveyed, including hand-crafted feature-based approaches and deep learning-based approaches.
\subsection{Plant Species Recognition}
\textbf{Hand-crafted descriptors}. Traditional hand-crafted descriptors manually extract leaf visual characteristics, such as leaf shape, texture, vein, and color, to identify plant species. Leaf shapes provide significant clues for botanists to identify species\cite{cope2012plant}. \cite{hu2012multiscale} proposed a novel contour-based shape descriptor to capture robust shape geometry which can be invariant to translation, rotation, scaling, and bilateral symmetry. \cite{wang2014bag} made the first attempt to introduce the idea of bag-of-words (BoW) for shape representation in which the shapes are decomposed into contour fragments. \cite{zhao2015plant} utilized a novel feature that captures global and local shape information independently. Furthermore, methods based on leaf texture and veins also have been presented in the past few decades. For example, \cite{cope2010plant} explored to use Gabor co-occurrences in plant texture classification. \cite{tang2015local} presented a texture description approach by combining LBP feature with a gray-level co-occurrence matrix (GLCM) for tea leaf classification. \cite{larese2014automatic} adopted a procedure for segmenting and classifying scanned legume leaves based only on the analysis of their veins. \cite{charters2014eagle} designed a new descriptor named Eagle which characterizes the overall venation structure using the edge patterns among neighboring regions. Recently, \cite{yang2021plant} put forward a feature fusion framework by integrating the shape information and the texture feature for plant leaf recognition.

\textbf{Deep learning methods}. In the past few years, several leaf identification approaches using deep learning have been developed. \cite{kumar2012leafsnap} designed a framework by first segmenting the leaf from the background, extracting features representing the curvature of the leaf's contour to identify numerous tree species. \cite{grinblat2016deep} attempted using a deep convolutional neural network (CNN) for the problem of plant identification from leaf vein patterns. \cite{lee2017deep} adopted a hybrid global-local leaf feature extraction method for plant classification. \cite{barre2017leafnet} developed a deep learning system to learn discriminative features from leaf images along with a classifier for species identification of plants. A procedure for segmenting and classifying scanned legume leaves based only on the analysis of their veins was proposed \cite{larese2014automatic}. \cite{shah2017leaf} presented a dual-path deep convolutional neural network (CNN) to learn joint feature representations for leaf images by exploiting their shape and texture characteristics. \cite{lee2018multi} proposed a multi-organ plant identification approach based on a CNN and recurrent neural network. They analyzed features of leaf and other organs, such as fruit, steam, or flower for plant species identification. \cite{beikmohammadi2022swp} put forward a framework for simulating botanist behaviors through three deep learning-based models. Nevertheless, these methods tend to perform poorly when transferred from species classification to cultivar classification. This is due to large intraclass distances and small inter-class distances among cultivars.

\subsection{Plant Cultivar Recognition}
Using leaf image patterns as clues for identifying plant species has achieved great success in the past decades. Recently, there is an increasing concern about whether leaf image patterns can also provide powerful discriminative information for cultivar-level recognition.

\textbf{Hand-crafted descriptors}
\cite{wang2020species} proposed a novel multi-scale sliding chord matching approach to extract leaf patterns that are distinctive for soybean cultivar identification.
\cite{yu2020patchy} put forward a novel Multi-Orientation Region Transform (MORT), which can effectively characterize both contour and structure features simultaneously. The proposed MORT can extract local structural features at various scales and orientations for comprehensive shape description.
\cite{zheng2019forest} introduced a leaf cultivar classification model based
on forest representation learning and multi-scale contour feature learning.
\cite{chen2022pairwise} designed a novel local binary pattern, named pairwise rotation-difference LBP (PRDLBP), for the characterization of leaf image patterns. 
\cite{chen2020invariant} attempted a new strategy of depicting leaf shapes by convolving the contour vector functions with Gaussian functions of different widths.
\cite{wang2020local} proposed a novel local R-symmetry co-occurrence method (RsCoM) for characterizing discriminative local symmetry patterns to distinguish subtle differences among cultivars.

\textbf{Deep learning methods}
\cite{zhang2021mfcis} combined leaf hand-crafted features and deep learning extracted features together to identify 5000 leaf images from 100 soybean cultivars. 
\cite{wang2022fusing} explored to fuse deep learning features of triplet leaves from different parts of soybean plants for effective cultivar recognition.
\cite{liu2020novel} proposed an efficient and convenient method for the classification of apple cultivars using a deep convolutional neural network, which is the delicate symmetry of human brain learning.
\cite{yu2021maskcov} integrated an auxiliary self-supervised learning module (MaskCOV) with a powerful in-image data augmentation scheme for cultivar classification.
\cite{yu2022spare} presented SPARE, a self-supervised part-erasing framework for ultra-fine-grained visual categorization. The key insight of SPARE is to learn discriminative representations by encoding a self-supervised module to predict the position of the erased parts.
\cite{yu2023mix} introduced a novel mixing attentive vision transformer (Mix-ViT) incorporated with a self-supervised module to address the UFGVC tasks.
\cite{tavakoli2021leaf} concluded that most work of plant classification is performed on the author’s dataset, which makes a comparison of different works difficult.
In order to achieve an adequate comparison among existing methods, we conduct comprehensive experiments with 22 competing methods on 7 benchmark datasets. Experimental results fully validate the contribution of our proposed method.
\begin{figure*}[t]
	\centering
	\centerline{\includegraphics[width=12cm]{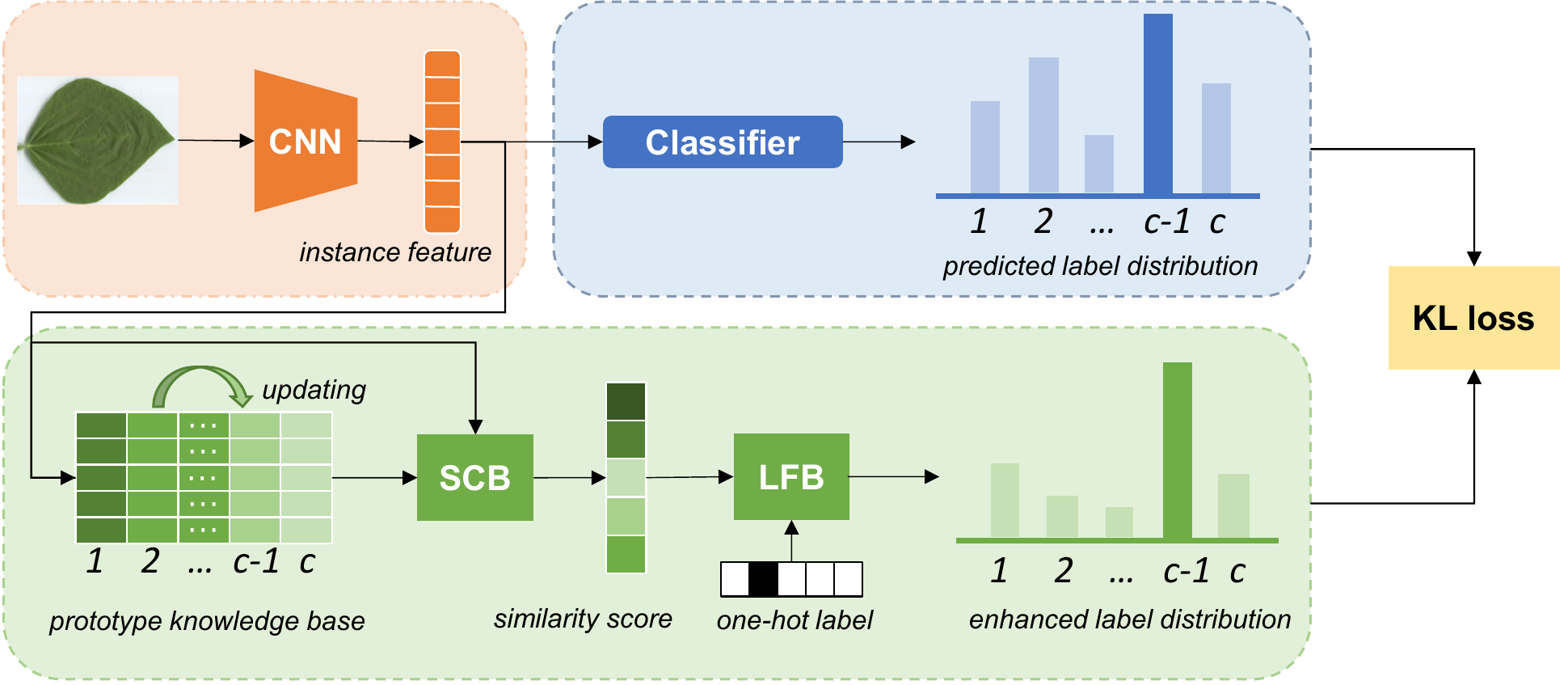}}
	\caption{Framework of the proposed Prototype-enhanced Learning(PEL) method. Given an input image, the CNN layers extract the instance feature representation. The classifier takes it to predict the instance label distribution. In parallel, we use the instance feature to update the prototype knowledge base, then compute the similarity score through Similarity Computing Block(SCB). The similarity score is then fused into the original one-hot label by Label Fusing Block(LFB), resulting in enhanced label distribution. Finally, the predicted label distribution and the enhanced label distribution are used to compute the Kullback-Leibler divergence loss.}
	\label{fig:framework}
\end{figure*}

\section{Methods}
In this section, we present the proposed Prototype-enhanced Learning (PEL) method. The PEL-based classification predictor consists of three modules: a basic feature encoder, a softmax classifier, and a label simulator. An overview of PEL is shown in figure \ref{fig:framework}.

\subsection{Basic Classification Predictor}
The PEL is designed to address the problem that conventional one-hot instance labeling makes the network overconfident and limits the model from learning semantic overlap between classes. We introduce PEL starting from the basic classification predictor. A basic classification predictor usually includes a feature encoder and a softmax classifier. Given the $i$-th input image $x_i$, the feature encoder extracts the instance feature $f(x_i)$ and feeds it to the classifier. The classifier is a single fully connected (FC) layer followed by a softmax layer to generate the predicted label distribution $\hat{y_i}$: 
\begin{equation}
\begin{split}
\hat{y_i} = softmax(FC(f(x_i))),
\end{split}
\label{eq:eqname1}
\end{equation}
\begin{equation}
\begin{split}
\hat{y_i}^j= \frac{exp(FC(f(x_i))_j/t_1)}{\sum_{n=1}^{N} exp(FC(f(x_i))_n/t_1)},
\end{split}
\label{eq:soft1}
\end{equation}
where $\hat{y_i}^j$ is the value in the $j$-th dimension of $\hat{y_i}$, $FC(f(x_i))_j$ is the $j$-th dimension of the predicted label logits from the $i$-th instance, $N$ is the number of categories, and $t_1$ is the temperature coefficient.

Traditional classification models apply cross-entropy loss between $\hat{y_i}$ and the ground-truth class $y_i$ to supervise the training process:
\begin{equation}
\begin{split}
CE(y_i, \hat{y_i}) = - \sum \limits_{j=1}^Ny_i^jlog(\hat{y_i}^j).
\end{split}
\label{eq:eqname1-1}
\end{equation}
In the above $y_i^j \in {\{0,1\}}$ specifies the one-hot ground-truth class distribution in the $j$-th dimension. 

However, one-hot labels are the same for samples of the same class, regardless of their contents. In fact, samples with the same label may have quite different contents, and naturally their label distributions should also be different. Although a theoretically realistic label distribution is not easy to achieve, we can try to model a distribution that reflects the degree of relationship between instances and labels.

\subsection{PEL-based Classification Predictor}
\begin{figure*}[t]
	\centering
	\centerline{\includegraphics[width=7cm]{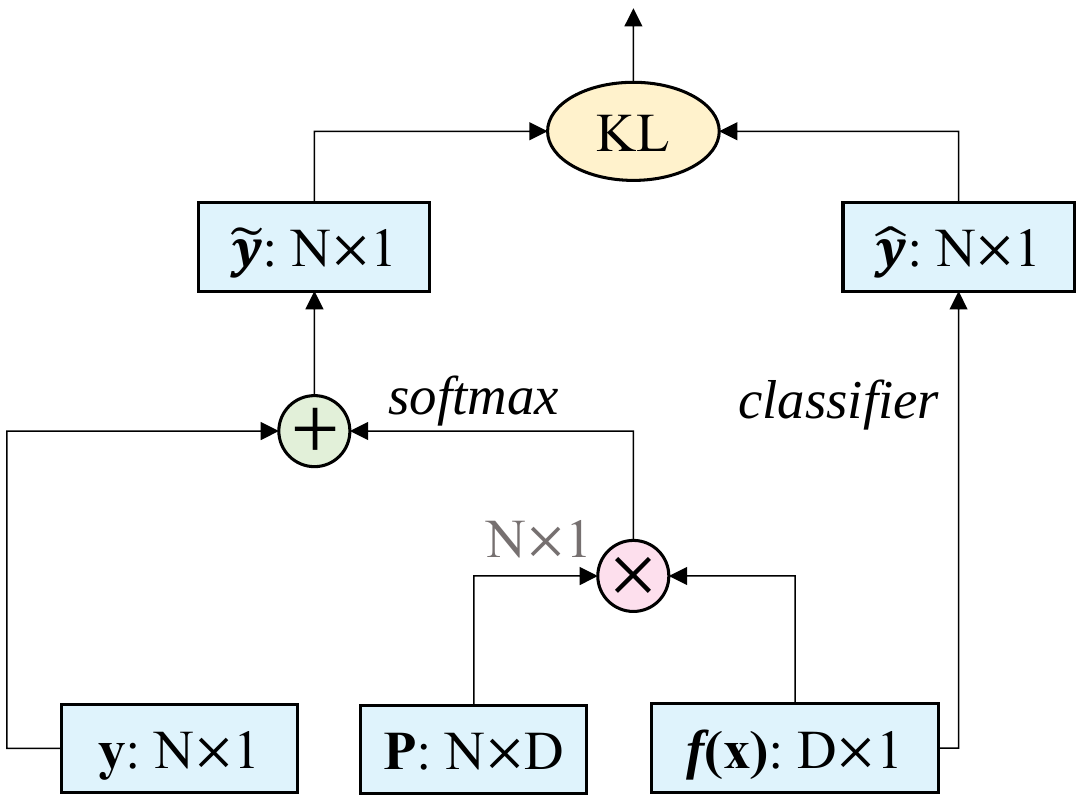}}
	\caption{ Pipeline of the proposed PEL method. In this picture, $\bigotimes$ is matrix multiplication, $\bigoplus$ is element-wise addition, and KL represents the Kullback-Leibler divergence loss.}
	\label{fig:process}
\end{figure*}

Different categories exhibit varying degrees of shared features with each other, implying that labels are not completely independent. With this basic intuition, we propose the PEL method with a label simulator to simulate the correlated label distribution. The pipeline of PEL is illustrated in Figure \ref{fig:process}. We first give a network initial prototypes \cite{snell2017prototypical} of each class as prior knowledge, which is termed prototypical knowledge base \bm{$\mathcal{P}$} $\in \mathbb{R} ^ {N \times D}$. We use normalized features generated from the final pooling layer for all backbone networks. This prototypical knowledge base is implemented as a matrix, where $N$ is the number of classes and $D$ is the prototype dimension. Each column represents a unique prototypical feature representation of a class. During training, normalized instance features in one batch will be grouped by their corresponding ground-truth labels. We then compute the mean feature representations in each group. The above process can be formulated as:
\begin{equation}
\begin{split}
\mathcal{F}_n = \frac {\sum \limits_{y_i \in n} f(x_i)} {|n|},
\end{split}
\label{eq:eqname2}
\end{equation}
where $x_i$ denotes the $i$-th instance in one batch, $y_i$ represents the ground-truth label of $x_i$, and $|n|$ represents the number of instances belonging to the $n$-th category, we have $\mathcal{F}_n$ denoting the mean feature representation of the $n$-th category.

The output mean features \bm{$\mathcal{F}$} are fed to the prototypical knowledge base \bm{$\mathcal{P}$} and the corresponding prototypes are updated accordingly by:
\begin{equation}
\begin{split}
\mathcal{P}_n \gets  \mathcal{P}_n + \alpha (\mathcal{F}_n - \mathcal{P}_n),
\end{split}
\label{eq:eqname3}
\end{equation}
where $\mathcal{P}_n$ denotes the prototypical feature representation of the $n$-th category, $\alpha \in (0,1)$ is the momentum coefficient that controls the updating rate of each prototype.

After that, the similarity computing block(SCB) will compute the similarity scores \bm{$w$} between each instance feature and the prototypical knowledge base. Intuitively, if an instance is close to a prototype, it may relate to the label of that prototype. Thus even if two instances belong to the same class, they carry different information w.r.t the similarity with other classes. The dependency among labels will be captured instance-specifically, which is superior to a uniform noise distribution as in label smoothing \cite{pereyra2017regularizing}. In practice, we use cosine similarity distance as the similarity metric. The similarity scores are then normalized by a softmax function. Given the input instance feature $f(x_i)$, the corresponding similarity score of the $i$-th instance $w_i$ is computed as:
\begin{equation}
\begin{split}
w_i =  softmax(\bm{\mathcal{P}} \times f(x_i)),
\end{split}
\label{eq:eqname4}
\end{equation}
\begin{equation}
\begin{split}
w_i^j= \frac{exp((\bm{\mathcal{P}} \times f(x_i))_j/t_2)}{\sum_{n=1}^{N} exp((\bm{\mathcal{P}} \times f(x_i))_n/t_2)},
\end{split}
\label{eq:soft2}
\end{equation}
where $w_i^j$ is the value in the $j$-th dimension of $w_i$, $(\bm{\mathcal{P}} \times f(x_i))_j$ is the $j$-th dimension of the predicted label logit from the $i$-th instance, $N$ is the number of categories, and $t_2$ is the temperature coefficient.

The label fusing block(LFB) takes the one-hot ground-truth targets \bm{$y$} and similarity scores \bm{$w$} as inputs, and fuses them with a weight coefficient. We define the enhanced label distribution $\widetilde{y_i}$ as:
\begin{equation}
\begin{split}
\widetilde{y_i} =  \beta y_i + w_i,
\end{split}
\label{eq:eqname5}
\end{equation}
where $\beta > 0$ is a weight coefficient to control the effect by the one-hot target $y_i$ of the $i$-th instance .

As a result, the enhanced label distribution not only contains groud truth information but also has inter-class similarity awareness. Finally, the enhanced label distribution takes place of the one-hot target to supervise the model training. The Kullback-Leibler divergence (KL) loss is applied to measure the distance $\mathcal{L}$ between the predicted and enhanced label distribution, which is formulated by:
\begin{equation}
\begin{split}
\mathcal{L} =  KL\mbox{-}divergence(\bm{\widetilde{y}}, \bm{\hat{y}}) \\
= \sum \limits_{i}^I \widetilde{y}_i log(\frac{\widetilde{y}_i}{\hat{y}_i}),
\end{split}
\label{eq:eqname6}
\end{equation}
where $I$ is the number of instances.

In the proposed PEL, the actual labels are dynamically changing along with the updating of prototypes. With the additional supervision from the prototypical knowledge base, models can learn much more information and reduce overfitting when facing small datasets. Moreover, the enhanced labels are more robust to mislabeled samples since the similarity scores can allocate probability to similar labels which usually include the right label. Please note that only the “CNN” and “classifier” are needed during inference, so the computational cost introduced is negligible.      
 
\section{Experiments}
\label{sec:experiments}
The performance of our proposed PEL is compared to 22 classification methods for a thorough evaluation. The benchmark methods are broadly split into two groups. One covers hand-crafted feature descriptors including SIT\cite{wang2016structure}, HSC\cite{wang2014hierarchical}, PH\cite{zhang2021mfcis}, RsCoM\cite{wang2020local} and MSCM\cite{wang2020species}. The second is deep learning methods consisting of 1) Basic CNN-based methods including Alexnet\cite{krizhevsky2017imagenet}, VGG-16\cite{simonyan2014very}, MobileNetV2\cite{sandler2018mobilenetv2} and Xception\cite{chollet2017xception}; 2) Basic transformer-based methods including ViT\cite{dosovitskiy2020image}, DeiT\cite{touvron2021training}, TransFG\cite{he2022transfg} and Hybrid-ViT\cite{dosovitskiy2020image}; 3) state-of-the-art methods for image recognition and FGVC including NTS-NET\cite{yang2018learning}, fast-MPN-COV\cite{li2018towards}, DCL\cite{chen2019destruction} and Cutmix\cite{yun2019cutmix}; 4) state-of-the-art methods for UFGVC including CF\cite{wang2022fusing}, MFCIS\cite{zhang2021mfcis}, MaskCOV\cite{yu2021maskcov}, SPARE\cite{yu2022spare}, Mix-ViT\cite{yu2023mix}.

\begin{table*}[h]
	\fontsize{8pt}{8pt}
	\selectfont
	\centering
	\begin{tabular}{cccc}
		\hline
		\hline
		{Dataset} & {Class} & {Train}& {Test}\\
		{Sweet cherry} & 88 & 3788 & 1623\\
		{CottonCultivar} & 80 & 240 & 240\\
		{SoyCultivarLocal} & 200 &600 &600  \\
		{SoyCultivarGene} & 1110 & 12763&11143 \\
		{SoyCultivarAge} & 198 &4950 &4950 \\
		{SoyCultivarGlobal} & 1938 &5814 & 5814 \\
		{SoyCultivar200} & 200 & 3000 &3000  \\
		\hline
		\hline
	\end{tabular}
	\caption{Statistics of the benchmark datasets.}
	\label{tab:dataset}
\end{table*}

\subsection{Leaf material and Benchmarks}
In this research, we adopted 7 benchmark datasets in our experimental evaluation. Table \ref{tab:dataset} summarizes the statistics for each dataset, i.e., the number of classes, training set, and testing set. For ultra-fine-grained visual classification (UFGVC), there are 7 image datasets including sweet cherry \cite{zhang2021mfcis}, SoyCultivarLocal \cite{yu2021benchmark}, CottonCultivar, SoyCultivarGlobal, SoyCultivarGene, SoyCultivarAge, and SoyCultivar200\cite{wang2020species}.

\begin{table*}[h]
	\fontsize{6pt}{6pt}
	\selectfont
	\centering
	\begin{tabular}{ccccccc}
		\hline
		\hline
		{\textbf{Methods}} & {Backbone} & {} & {} & {\textbf{Accuracy(\%)}} & {} & {}\\
		\cline{3-7}
		{} & {} & {Cotton} & {S.Loc} & {S.Gene} & {S.Age} & {S.Glo}\\
		\cline{3-7}
		{\emph{Alexnet}\cite{krizhevsky2017imagenet}} & Alexnet  & 22.92  & 19.50  & 13.12  & 44.93  & 13.21 \\
		{\emph{VGG-16}\cite{simonyan2014very}} & VGG-16  & 50.83  & 39.33  & 63.54  & 70.44  & 45.17 \\
		{\emph{MobileNetV2}\cite{sandler2018mobilenetv2}} & ResNet-50  & 49.58  & 34.67  & 63.17  & - & 31.66 \\
		{\emph{NTS-NET}\cite{yang2018learning}}& ResNet-50  & 51.67  & 42.67  & -  & -  & - \\
		{\emph{fast-MPN-COV}\cite{li2018towards}} & ResNet-50  & 50.00  & 38.17  & 45.26  & -  & 11.39 \\
		{\emph{Cutmix}\cite{yun2019cutmix}} & ResNet-50  & 45.00  & 26.33  & 66.39  & 62.68  & 30.31 \\
		{\emph{DCL}\cite{chen2019destruction}} & ResNet-50 & 53.75  & 45.33  & 71.41  & 73.19  & 42.21\\
		{\emph{MaskCOV}\cite{yu2021maskcov}} & ResNet-50  & 58.75 & 46.17  & 73.57  & 75.86  & 50.28\\
		{\emph{SPARE}\cite{yu2022spare}} & ResNet-50  & 60.42  & 44.67  & 79.41  & 75.72  & {56.45 }\\
		{\emph{ViT}\cite{dosovitskiy2020image}} & Transformer  & 52.50  & 38.83  & 53.63  & 66.95  & {40.57 }\\
		{\emph{DeiT}\cite{touvron2021training}} & Transformer  & 54.17  & 38.67  & 66.80  & 69.54  & {45.34 }\\
		{\emph{TransFG}\cite{he2022transfg}} & Transformer  & 54.58  & 40.67  & 22.38  & 72.16  & {21.24 }\\
		{\emph{Hybrid-ViT}\cite{dosovitskiy2020image}} &  Transformer\&ResNet  & 50.83  & 37.00 & 71.74  & 73.56 & {18.82 }\\
		{\emph{Mix-ViT}\cite{yu2023mix}} &  Transformer\&ResNet  & 60.42  & 56.17 & \underline{79.94}  & 76.30  & {51.00 }\\
		\textbf{\emph{PEL}} &  ResNet-50  & \underline{62.92} & \underline{58.67} & 79.50  & \underline{81.45}  & \underline{57.58}\\
		\textbf{\emph{PEL}} & DenseNet-121 & \textbf{63.33} & \textbf{59.33} & \textbf{81.49}  & \textbf{82.30}  & \textbf{60.56}\\
		\hline
		\hline
	\end{tabular}
	\caption{The top 1 classification accuracy on the CottonCultivar (Cotton), SoyCultivarLocal (S.Loc), SoyCultivarGene (S.Gene), SoyCultivarAge (S.Age) and SoyCultivarGlobal (S.Glo) datasets. Results style: \textbf{best} and \underline{second-best} among each method.}
	\label{tab:soyall}
\end{table*}

\begin{table*}[h]
	\fontsize{6.5pt}{6.5pt}
	\selectfont
	\centering
	\begin{tabular}{cccccccc}
		\hline
		\hline
		{\textbf{Methods}} & {Backbone} & {} & {} & {\textbf{Acc(\%)}} & {} & {} & {}\\
		\cline{3-8}
		{} & {} & {R1} & {R3} & {R4} & {R5} & {R6} & {Avg}\\
		\cline{3-8}
		{\emph{Alexnet}\cite{krizhevsky2017imagenet}} & Alexnet  & 49.90  & 44.65  & 45.15  & 47.47  & 37.47 & 44.93 \\
		{\emph{VGG-16}\cite{simonyan2014very}} & VGG-16  & 72.32  & 72.53  & 74.95  & 71.11  & 61.31 & 70.44\\
		{\emph{NTS-NET}\cite{yang2018learning}} & ResNet-50  & 63.94  & 67.68  & 51.52  & 61.41  & 55.76 & 60.06\\
		{\emph{fast-MPN-COV}\cite{li2018towards}} & ResNet-50  & 67.68  & 64.55  & 66.87  & 68.49  & 50.71 & 63.66 \\
		{\emph{Cutmix}\cite{yun2019cutmix}} & ResNet-50  & 65.56  & 59.19  &64.24  & 68.79  & 53.64 & 62.28 \\
		{\emph{DCL}\cite{chen2019destruction}} & ResNet-50 & 76.87  & 73.84  & 76.16  & 76.16  & 62.93& 73.19\\
		{\emph{MaskCOV}\cite{yu2021maskcov}} & ResNet-50  & 79.80 & 74.65  & 79.60  & 78.28  & 66.97& {75.86}\\
		{\emph{SPARE}\cite{yu2022spare}} & ResNet-50  & 78.28  & 79.90  & 78.69  & 77.27  & {64.44}& {75.72}\\
		{\emph{ViT}\cite{dosovitskiy2020image}} & Transformer  & 69.29  & 64.55  & 70.40  & 71.01  & {59.49}& {66.95}\\
		{\emph{DeiT}\cite{touvron2021training}} & Transformer  & 73.03  & 70.40  & 69.09  & 74.65  & {60.51}& {69.54}\\
		{\emph{TransFG}\cite{he2022transfg}} & Transformer  & 74.95  & 74.55  & 74.24  & 76.26  & {60.81}& {72.16}\\
		{\emph{Hybrid-ViT}\cite{dosovitskiy2020image}} &  Transformer\&ResNet  & 77.17  & 76.97 & 74.75  & 76.36 & 63.53 & {73.56}\\
		{\emph{Mix-ViT}\cite{yu2023mix}} &  Transformer\&ResNet  & 79.29  & 77.17 & 77.98  & 79.19  & {67.88}& {76.30}\\
		\textbf{\emph{PEL}} &  ResNet-50  & \underline{84.24} & \underline{82.53} & \underline{84.14}  & \textbf{84.64}  & \textbf{71.71}& \underline{81.45}\\
		\textbf{\emph{PEL}} &  DenseNet-121  & \textbf{85.85} & \textbf{83.53} & \textbf{86.26}  & \underline{84.44}  & \underline{71.41}&  \textbf{82.30}\\
		\hline
		\hline
	\end{tabular}
	\caption{The top 1 classification accuracy on the five subsets R1, R3, R4, R5 and R6 of the SoyCultivarAge dataset. “Avg” denotes the average classification accuracy of the five subsets. Results style: \textbf{best} and \underline{second-best} among each method.}
	\label{tab:soyage}
\end{table*}

\begin{table*}[h]
	\fontsize{8pt}{8pt}
	\selectfont
	\centering
	\begin{tabular}{cccccc}
		\hline
		\hline
		{\textbf{Methods}} & {Backbone} & {}  & {\textbf{Accuracy(\%)}} & {} & {}\\
		\cline{3-6}
		{} & {} & {Low} & {Mid} & {Up} & {Avg} \\
		\cline{3-6}
		{\emph{SIT}\cite{wang2016structure}} & -  & 18.30  & 12.15  & 13.20  & 14.55 \\
		{\emph{HSC}\cite{wang2014hierarchical}} & -  & 23.00  & 18.80  & 16.15  & 19.32 \\
		{\emph{RsCoM}\cite{wang2020local}} & - & 30.15  & 28.04  & 31.15  &  29.78 \\
		{\emph{MSCM}\cite{wang2020species}} & -  & 34.70 & 33.55  & 31.03  &  33.09 \\
		{\emph{CF}\cite{wang2022fusing}} & ResNet-50  & 37.40  & 40.10  & 39.01  &  38.84 \\
		{\emph{CF}\cite{wang2022fusing}} & DenseNet-121  & 43.50  & 47.15  & 44.90  &  45.18 \\
		{\emph{MFCIS}\cite{zhang2021mfcis}} & Xception  & 76.00   & 76.02 &79.67& 77.23\\
		{\emph{DCL\cite{chen2019destruction}}} & ResNet-50  & 79.52 & \underline{85.40} & \underline{87.20}  & \underline{84.04} \\
		{\emph{MaskCOV}\cite{yu2021maskcov}} & ResNet-50  & 79.70 & 81.20 & 83.50  &  81.47\\
		{\emph{Mix-ViT}\cite{yu2023mix}} &  Transformer\&ResNet  &78.82   & 81.74 & 84.13  & 81.56 \\
		\textbf{\emph{PEL}} &  ResNet-50  & \underline{79.90} & {84.50} & 85.41  & 83.27 \\
		\textbf{\emph{PEL}} &  DenseNet-121  & \textbf{80.55} & \textbf{86.83} & \textbf{88.84}  & \textbf{85.34} \\
		\hline
		\hline
	\end{tabular}
	\caption{The top 1 classification accuracy on the three subsets Low, Mid and Up of the SoyCultivar200 dataset. “Avg” denotes the average classification accuracy of the three subsets. Results style: \textbf{best} and \underline{second-best} among each method.}
	\label{tab:soy200}
\end{table*}

\begin{table*}[h]
	\fontsize{8pt}{8pt}
	\selectfont
	\centering
	\begin{tabular}{cccc}
		\hline
		\hline
		{\textbf{Methods}} & {Backbone} & {\textbf{Accuracy(\%)}} &{}\\
		{} & & {Cherry} & {SoyCultivar200} \\
		\cline{3-4}
		{\emph{HSC}\cite{wang2014hierarchical}} & -  & 16.47  & 45.45 \\
		{\emph{PH}\cite{zhang2021mfcis}} & - & 42.08  & - \\
		{\emph{DCNN}\cite{liu2020novel}} & - & 32.55  & - \\
		{\emph{RsCoM}\cite{wang2020local}} & - & -  & 64.93  \\
		{\emph{MSCM}\cite{wang2020species}} & -  & - & 72.40 \\
		{\emph{CF}\cite{wang2022fusing}} & DenseNet-121  & -  & 83.55  \\
		{\emph{Xception}\cite{chollet2017xception}} & Xception  & 66.52  & - \\
		{\emph{MFCIS}\cite{zhang2021mfcis}} & Xception  & 83.52   & 78.00 \\
		{\emph{DCL}\cite{chen2019destruction}} & ResNet-50 & \underline{95.50} & 87.50 \\
		{\emph{MaskCOV}\cite{yu2021maskcov}} & ResNet-50  & 91.75  & 83.92\\
		{\emph{Mix-ViT}\cite{yu2023mix}} &  Transformer\&ResNet  &  93.10 & 84.50\\
		\textbf{\emph{PEL}} &  ResNet-50  & {95.34}   & \textbf{89.33}\\
		\textbf{\emph{PEL}} &  DenseNet-121  & \textbf{95.62}  &  \underline{89.20}\\
		\hline
		\hline
	\end{tabular}
	\caption{The top 1 classification accuracy on the sweet cherry and SoyCultivar200 dataset. Results style: \textbf{best} and \underline{second-best} among each method.}
	\label{tab:cherry}
\end{table*}

\subsection{Implementation Details}
We implement our proposed PEL in Pytorch. All networks are trained and tested on a single Tesla A-100 GPU. We evaluate PEL on two widely used backbones: ResNet-50 and DenseNet-121. They are initialized by the ImageNet pre-trained model. The input images are resized to 512$\times$512 and center cropped into 448$\times$448. Random rotation with a degree of 15 and random horizontal flips are adopted for data augmentation. The above are standard setups in the literature. We adopt an SGD optimizer with a momentum of 0.9 and weight decay 1e-4. The base learning rate is 0.001 and the batch size is set to 8 for all datasets, except for SoyCultivarGlobal with a learning rate of 0.01 and a batch size of 64. The number of epochs is selected according to the data size. All datasets use epoch 120 except for the larger dataset SoyCultivarGene with epoch 200. The temperature coefficients $t1$ (Eq. \ref{eq:soft1}) and $t2$ (Eq. \ref{eq:soft2}) are both set as 1. The momentum coefficient $\alpha$ (Eq. \ref{eq:eqname3}) is empirically set to 0.9. Besides, we set the weight coefficient $\beta$ (Eq. \ref{eq:eqname6}) as 6 in all datasets except for SoyCultivarGene and SoyCultivarGlobal with $\beta$ as 8. Ablation studies on the effect of the weight coefficient are reported in section \ref*{sec:ablation}. Given an ImageNet-pretrained backbone network, the prototypical knowledge base is initialized as the mean feature of each category on the training set. During test time, the architectures of SCB and LFB are removed.

\subsection{Performance Comparison}
To assess the capability of PEL in the UFGVC tasks, we conduct evaluations on 7 ultra-fine-grained datasets. Most of the datasets have been split into train and test sets, except in the sweet cherry dataset, we adopt ten random train and test splits as in MFCIS\cite{zhang2021mfcis} with a split ratio of 7:3. 

\textbf{Comparison on CottonCultivar Dataset.} As reported in Table \ref{tab:soyall}, the proposed PEL surpasses all the competitive methods w.r.t the classification accuracy. Specifically, PEL based on DenseNet-121 achieves the best performance, which is 2.91\% higher than the second-best methods Mix-Vit and SPARE.

\textbf{Comparison on SoyCultivarLocal Dataset.} PEL presents favorable performance results as listed in Table \ref{tab:soyall}. DenseNet-121 based PEL achieves the best performance at 59.33\% among 15 outstanding methods, followed by the recently proposed state-of-the-art method Mix-Vit with a classification accuracy of 56.17\%. 

\textbf{Comparison on SoyCultivarGene Dataset.} In Table \ref{tab:soyall}, DenseNet-121 based PEL obtains the highest classification accuracy of 81.49\%, outperforming all the other 14 methods. We observe that Mix-ViT shows its superiority while evaluated on a relatively larger dataset (SoyCultivarGene) with more than 20 thousand image samples.

\textbf{Comparison on SoyCultivarGlobal Dataset.} As shown in Table \ref{tab:soyall}, PEL are competitive compared to the state-of-the-art methods. For example, DenseNet-121 based PEL gains 4.11\% improvement over the second-best approach SPARE.

\textbf{Comparison on SoyCultivarAge Dataset.} SoyCultivarAge dataset covers five subsets where each subset contains leaves collected from a specific cultivar period. The comparative performance of 14 competing methods is summarized in Table \ref{tab:soyage}. For each subset, PEL shows dominant capability compared to other approaches. Particularly, DenseNet-121 based PEL achieves the best performance with a significant margin of 6.05\% over the second-best method MaskCOV in the R1 period subset. Moreover, we observe that R6 exhibits a substantially more than 10\% lower accuracy than other periods. 

\textbf{Comparison on SoyCultivar200 Dataset.} Table \ref{tab:soy200} presents the performance of existing hand-crafted feature extraction methods as well as competitive deep learning methods on the subsets of the SoyCultivar200 dataset. According to the results, deep learning methods are much superior to hand-crafted feature extraction methods. Furthermore, results demonstrate that soy leaves collected from the low part of one plant are less discriminative than those from the mid and up parts. The finding suggests that the newly emerging leaves in the upper part of a plant carry richer information than the mature leaves in the lower part. In addition, we mix up all subsets in SoyCultivar200, train and test each approach on the whole dataset. As Table \ref{tab:cherry} shows, PEL achieves the best performance 89.33\%.

\textbf{Comparison on sweet cherry Dataset.} In Table \ref{tab:cherry}, we compare PEL with the existing methods on the sweet cherry dataset. We observe that DenseNet-121 based PEL achieves encouraging classification accuracy which is 95.62\%. Besides, DCL also shows its advantage with only 0.12\% lower than our method.

The overall performance is consistent with the number of samples in each category. There are only around 3 samples per cultivar in CottonCultivar, SoyCultivarLocal, and SoyCultivarGlobal, which involves a few-shot problem in the UFGVC tasks. Therefore, the performances on these three benchmark datasets are relatively low which are around 60\%. As we know that few shot datasets are prone to overfitting, prototypical learning is a conventional way for few shot tasks. Thus, our prototype-enhanced learning approach helps alleviate the few-shot problem in UFGVC tasks. Concluded from Table \ref{tab:soy200} and Table \ref{tab:cherry}, deep learning methods have a very high performance ceiling that dominates comparison with hand-crafted feature extraction methods. In short, the results demonstrate the potential of our proposed method toward narrowing the performance gap between the FGVC and UFGVC tasks.

\subsection{Ablation Studies}
\label{sec:ablation}
\begin{figure*}[t]
	\centering
	\centerline{\includegraphics[width=12cm]{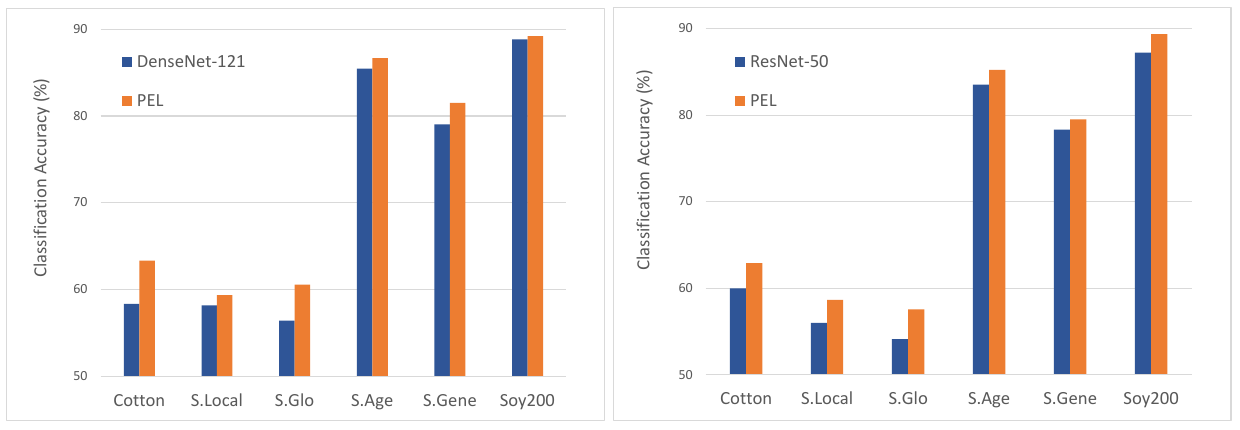}}
	\caption{Performance comparison between baselines and PEL.}
	\label{fig:ablation}
\end{figure*}

\begin{table*}[h]
	\fontsize{8pt}{8pt}
	\selectfont
	\centering
	\begin{tabular}{cccc}
		\hline
		\hline
		{\textbf{Backbone}} & \textbf{Dataset} & \textbf{weight} & {\textbf{Accuracy(\%)}}\\
		\hline
		{} &   & 4  & 86.46 \\
		{\emph{ResNet-50}\cite{he2016deep}} & {SoyCultivar200} & 6 &  \textbf{89.33}\\
		{} & {} & 8 & 88.20\\
		\hline
		{} &   & 6  & 44.94 \\
		{\emph{DenseNet-121}\cite{huang2017densely}} & SoyCultivarGlobal & 8  &  \textbf{60.56} \\
		{} & {} & 10 & 59.49\\
		\hline
		\hline
	\end{tabular}
	\caption{Ablation on the effect of weight coefficeint regarding classification accuracy on two benchmark datasets.}
	\label{tab:ablation}
\end{table*}

\begin{table*}[h]
	\fontsize{8pt}{8pt}
	\selectfont
	\centering
	\begin{tabular}{ccccc}
		\hline
		\hline
		{\textbf{Method}} & \textbf{Backbone} & \textbf{Dimension} & {\textbf{Parameters}} & {\textbf{GFLOPs}}\\
		\hline
		
		{ViT}\cite{dosovitskiy2020image} & {Transformer} & \textbf{0.7K} & 86.24M & -\\
		{DeiT}\cite{touvron2021training} & {Transformer} & \textbf{0.7K} & 86.24M & -\\
		{TransFG}\cite{he2022transfg} & {Transformer} & \textbf{0.7K} &  86.80M & -\\
		{Hybrid ViT}\cite{dosovitskiy2020image} & {Trans.\&ResNet} & \textbf{0.7K} & 99.20M  & -\\
		{Mix-ViT}\cite{yu2023mix} & {Trans.\&ResNet} &  \textbf{0.7K} & \textbf{35.42M} & -\\
		\hline
		{Alexnet}\cite{krizhevsky2017imagenet} & {Alexnet} & 4K & 61.10M & \textbf{2.84}\\
		{VGG-16}\cite{simonyan2014very} & {VGG-16} &4K  & 138.36M & 61.64\\
		{ResNet-50}\cite{he2016deep} & {ResNet-50} &2K  &25.56M  & 16.47\\
		{DenseNet-121}\cite{huang2017densely} & {DenseNet-121} & \textbf{1K} & \textbf{7.0M} & {11.53}\\
		{DCL}\cite{chen2019destruction} & {ResNet-50} &2K  & 23.68M & 16.47 \\
		{MaskCOV}\cite{yu2021maskcov} & {ResNet-50}  &2K  & 23.75M & 16.47\\
		\textbf{PEL} & {ResNet-50} & 2K & 25.56M & 16.47\\
		\textbf{PEL} & {DenseNet-121} & \textbf{1K}  & \textbf{7.0M} & {11.53} \\
		\hline
		\hline
	\end{tabular}
	\caption{Performance comparison with respect to feature dimension, parameters and GFLOPs with input size of 448$\times$448. The results in bold represent the most efficient in each group of methods.}
	\label{tab:cost}
\end{table*}

We conduct a comprehensive ablation study to further verify the contribution of the proposed PEL method. The baseline methods includes ResNet-50\cite{he2016deep} and DenseNet-121\cite{huang2017densely}. As shown in Figure \ref{fig:ablation}, PEL consistently improves the performance over baselines, indicating the effectiveness of the PEL method. 

\textbf{Effect of weight coefficient on PEL.} As we mentioned before, the similarity scores conducted in PEL may introduce noises in the early training stage when prototypes are unstable and imprecise. Therefore, we implement ablation studies of the effect of the weight coefficient $\beta$ (Eq. \ref{eq:eqname6}) on SoyCultivar200 and SoyCltivarGlobal respectively. In Table \ref{tab:ablation}, we compare ResNet-50 based PEL evaluated on SoyCultivar200 with different weight coefficients from 4 to 8. We set $\beta$ to 6 for experiments on all datasets except SoyCultivarGene and SoyCultivarGlobal. As presented in Table \ref{tab:ablation}, the weight coefficient of 8 achieves the best performance by the DenseNet-121 based PEL on SoyCltivarGlobal. We set $\beta$ as 8 for all experiments on SoyCultivarGene and SoyCultivarGlobal.

\textbf{Computational cost.} In Table \ref{tab:cost}, we list the comparison results in three aspects: feature dimension (a key measurement in classification tasks), computational complexity (GFLOPs), and memory consumption (model parameters). For a fair comparison, the input size is set to 448$\times$448 and the category number is set to 80 for all the competing methods. The DenseNet-121 based PEL is relatively effective in each aspect with only 11.53 GFLOPs, 1K feature dimension, and 7M model parameters.

\section{Conclusions}
\label{sec:conclusions}
In this paper, we propose a novel PEL method for the challenging ultra-fine-grained image recognition. As an open topic, cultivar identification covers three key issues: high similarity among classes, label mislabeling, and lack of data. On top of that, PEL generates enhanced label distributions that not only contain the target category but also have inter-class similarity awareness, thus being more sensitive to the similarity degree among categories. Moreover, the enhanced labels are more robust to mislabeled samples since the similarity scores can allocate probability to similar labels which usually include the right label, thus the model can still learn useful information from mislabeled samples. With the additional supervision from the prototypical knowledge base, models can reduce overfitting when facing small datasets. Our method has achieved superior performance on 7 UFGVC benchmark datasets compared to 22 competitive methods. 

The beneficial performance and excellent efficiency confirm the superiority of PEL in the UFGVC tasks. However, results demonstrate that on datasets lacking training samples, such as CottonCultivar, SoyCultivarLocal, and SoyCultivarAge, the overall performance remains around 60\%, and there is still much room for improvement. Therefore, the lack of data is still a crucial challenge in UFGVC tasks. Since UFGVC tasks are often accompanied by the few-shot problem, effective methods to avoid overfitting in UGFVC can be a key research point in future work.

\section{Acknowledgement}
The work was supported in part by the National Science Foundation of China (62106236) and Research Program of Zhejiang Lab (2021PE0AC04 and 2021PE0AC05).
\bibliography{refs}

\end{document}